# AI Powered Anti-Cyber Bullying System using Machine Learning Algorithm of Multinomial Naïve Bayes and Optimized Linear Support Vector Machine

Interception of Cyberbully Contents in a Messaging System by Machine Learning Algorithm

Tosin Ige[1]

Department of Computer Science
University of Texas at El Paso, Texas, USA

Sikiru Adewale[2]

Department of Computer Science
Virginia Technological University, Virginia, USA

*Abstract*—**"Unless and until our society recognizes cyber bullying for what it is, the suffering of thousands of silent victims will continue." ~ Anna Maria Chavez. There had been series of research on cyber bullying which are unable to provide reliable solution to cyber bullying. In this research work, we were able to provide a permanent solution to this by developing a model capable of detecting and intercepting bullying incoming and outgoing messages with 92% accuracy. We also developed a chatbot automation messaging system to test our model leading to the development of Artificial Intelligence powered anti-cyber bullying system using machine learning algorithm of Multinomial Naïve Bayes (MNB) and optimized linear Support Vector Machine (SVM). Our model is able to detect and intercept bullying outgoing and incoming bullying messages and take immediate action.**

## I. Introduction

Hatred, violence, and hostility in modern world can take several form [4],[5],[6],[2], one of which is cyber bullying using modern day technology medium. While the era of internet had brought in tremendous innovation and improvements to our daily activities and overall way of life, it had also opened floodgates for cyber bullying. Any devastating act in the mode of aggressive or abusive behavior toward people regarding digital interactions is cyber bullying [1] Fig. 1

The impact of social media like Instagram, Facebook, Twitter, WhatsApp, etc. on daily basis cannot be over emphasize as they had greatly influence modern way of communication As useful as social media is, it is a medium for promoting hatred, harassment, racism, etc. which is currently affecting millions of people across the globe.

Statistical record from 2019 Cyber bullying Data shows that 95% of teens in the U.S. are online, and the vast majority has access to internet on their mobile device, makes social media platform the most common medium for cyber bullying [11]. About 37% of young people between the ages of 12 and 17 have been bullied online. 30% have had it happen more than once [8],[9],[10]. The impact of cyber bullying is very visible in the world today as it had result in several hatred, trauma, depression, and untimely death. Ryan Halligan (1989–2003), age 13, was an American student from Essex Junction, Vermont, who died by suicide at the age of 13 after allegedly being bullied by his classmates in person and online. While Jeff Weise (1988–2005), age 16, who was also an American high school student who committed the Red Lake shootings killing nine people and himself by suicide after being severally attack by cyber bullying Fig. 2.

There had been several measures and implementations put in place to prevent cyber bullying as a solution but none of them have actually solve cyber bully as the effect of cyber bullying is still obvious in our society.

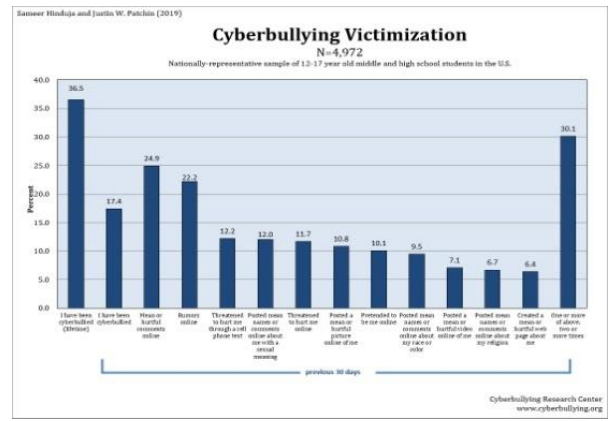

Fig. 1. Statistical Analysis and Victim of Cyberbullying.

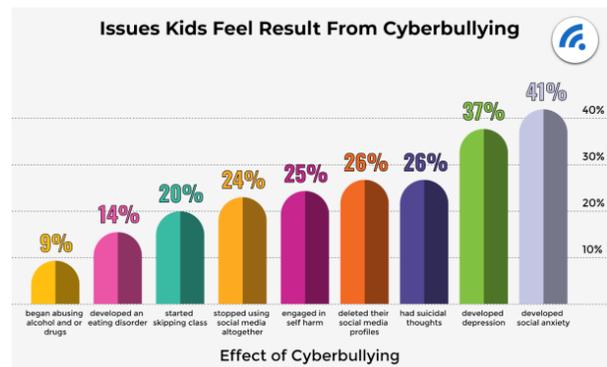

Fig. 2. Statistical Effects of Cyberbullying.

In this research work, Multinomial Naïve Bayes and Optimized Linear Support Vector Machine of machine learning algorithm was successfully used to implement an Artificial Intelligent Powered anti-cyber bullying system capable of detecting and intercepting online cyberbullying messages and filtered [3] them so that the intending recipient doesn't receive it. To ensure a better coverage across billions of device worldwide, we deployed our model and exposed it to a web application programming interface (API). The REST API was developed with Flask python framework, so the API load our deployed model, call it with parameterized input, and then indicated the status if it is cyberbullying or not using natural language processing capability of machine learning for detection and interception. All that any social media website like Facebook, Twitter, Instagram, Linkedin, Snapchat, YouTube, WhatsApp, etc. needed to do is to integrate it on their platform by calling the API alone. With this implementation, the problem of cyber bullying had been laid to rest. In order to test our implementation in real time, we developed a chatbot automation messaging system, and then use it to consume our restful API service, we got an accuracy of 92% and the model was able to automatically detect and intercept any outgoing and incoming bullying message and then ensure that the intending receiver doesn't receive the message. We also estimate the time taken by the restful API service to detect, intercept, and process messages and discovered that it is negligible fraction of seconds which simply that our implementation can be implemented, deployed, and use in real time.

## II. BACKGROUND STUDY

Over the years, there had been several efforts to address the issue of cyber bullying, but none had actually enforce the prevention which is why the rate of cyberbully had remain so high in our society. The authors in [14] from the MIT Media Lab, led by KarthikDinakar implemented an algorithm based on clustering and classification that is able to detect and categorize group of words in an online interaction [3]. The algorithm correctly categorizes contents in online interaction with high accuracy. It could not be used to combat cyber bullying both in real time. Several models and algorithm had been developed from various research labs, but none had been able to combat cyber bullying in real time. Logical probabilistic model of approach was used to develop a socio-linguistic model capable of detecting cybeybullying and the role play in the context of the conversation [15], Natural Language processing method of approach had been employed to identify cyber bullying on chat conversation [11], while Raisi uses Co-trained ensembles for weakly supervised bullying detection of embedded models used RNN and node2vec learners for detecting harassment and bullying from text base data [7].

As different research had been carried out from different research labs and centers, many of which involves the use of Different machine and deep learning techniques to catch word phrases and slangs like k Nearest Neighbor (kNN), Linear Regression (LR), Random Forests (RF), Logistic Regression (LogR), Boosting (Bos), Bagging (Bgg), Adaboost (ADB), Multiple Regression (MR), Maximum Entropy (MaxE), etc.

are being used for detection of cyberbullying on social media [12],[13].

Kainat et al, in 2021 proposed machine learning based algorithm which detects harassment actively and alert user to take action against it [13] But Kainat et al proposal is not feasible in real time and does not solve the problem of cyberbully due to the following reasons:

*1)* It does not intercept the incoming message: A real time anti cyberbully must intercept the incoming message before getting to the receiver.

*2)* The receiver have to take action to delete the content: If someone harasses me online, even though I delete the message, the fact that I already receive it will have psychological effect on me.

*3)* Reliance on csvfile from both sender and receiver which is not feasible due to cost and scalability.

*4)* No API exposure for wide coverage.

In this research work, we were able to implement anti-cyberbully system, which is able to automatically intercept an incoming message before getting to the receiver and take necessary action. Three major steps were involved:

*1)* We used Multinomial Naïve Bayes (MNB) and Optimized Linear Support vector Machine (svm) to train our model.

*2)* Deployed the model.

*3)* Build a Restful API service using Flask framework of Python.

## III. RESEARCH METHODOLOGY

For this research, we used different sources of dataset collection which are related to cyber-bullying. The data is from different social media platforms like Kaggle, Twitter, Wikipedia Talk pages and YouTube. The data contain text and labeled as bullying or not. The data contains different types of cyber-bullying like racism, hate speech, aggression, insults and toxicity. Having set up our cloud environment and the necessary programs written, we proceeded by adopting the standard machine learning data cleansing and preparation techniques Fig. 3.

Data cleaning: We use preprocessing and cleaning methods to remove incomplete data that might cause system failure and also affect output prediction. Rows containing missing values where completely removed. We also use different methods to identify and remove noisy data, outliners, and other factors which can influence the output result.

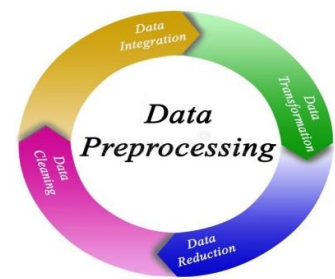

Fig. 3. Data Preprocessing Steps.

Data Reduction for Data Quality: In order to maintain data integrity, we needed to deal with all rows containing null value; hence we opted for python library tool called PyCaret. We have two options which are either to automatically fill all the null values or to remove any row(s) containing null values weighted. Having weighted the risk involves in both, we decided to remove any row with null or empty value from the data, and this was done using PyCaret python library.

Data Transformation: For us to make our data to acceptable format for easy data mining and pattern recognition, it needed to undergo some data transformation. To ensure data is fully transformed to acceptable format, we used Discretization, normalization, and data aggregation technique.

Data Mining: Having successfully set up our apparatus which includes IBM cloud object storage facility, running background window service, active web service, scheduler, and with the data being thoroughly pre-processed and transformed.

Unlike current data mining in cloud computing process in which, one needs to make direct connect to the data warehouse or called the csv file. We only called our web API which was developed and hosted on the cloud as seen below;

Our Artificial intelligence powered anti-cyberbullying system is based on the system of intercepting an outgoing message from the sender. It automatically intercept an incoming message before getting to the receiver and validate the status whether it is bullying or non-bullying. If it is a bullying message, it is automatically blocked from getting to the receiver, and the status of the message will be displayed as 'not delivered' to the sender using a natural language processing mechanism of artificial intelligence. We use machine learning algorithm of Multinomial Naïve Bayes (MNB) and Linear Support Vector Machine (SVM) to train our model, and then expose it to a restful API for global coverage.

### A. Implementation with Multinomial Naïve Bayes

Multinomial Naïve Bayes is based on the binomial distribution which is derived from series of combinatorial theorem (Fig. 4).

P (Xi|Xj) = P (Xi) for any distinct Xi and Xj as we as P (Xi|X) = P (Xi) for any x⊂x\xi.

Expansion and product simplification of the chain rule:

P (Ck,X1, . . . , Xn) = P (Ck) πni=1 P(X1|Ck)

In order to prevent the web service from being overwhelm due to multiple calling of the endpoint at regular interval, we developed a background window service to support the restful API service. Both the web API and the window service are picking records from the integrated data warehouse and pushing to the CSV file on the IBM Object Cloud Storage facility. They automatically pick new records to the CSV, and if a record is modified, it will be picked and modified on the CSV as well. The essence of the scheduler which was written in python is to be calling the web API at regular interval to check and push from the integrated data warehouse to the CSV file.

With the successful setting up of cloud environment and the necessary software programs being in fully execution, we proceeded by adopting the following data cleansing and preparation techniques.

Data cleaning: We use preprocessing and cleaning methods to remove incomplete data that might cause system failure and also affect output prediction. Rows contain- ing missing values where completely removed. We also use different methods to identify and remove noisy data, outliners, and other factors which can influence the output result.

Data Reduction for Data Quality: In order to maintain data integrity, we needed to deal with all rows containing null value; hence we opted for python library tool called PyCaret. We have two options which is either to automatically fill all the null values or to remove any row(s) containing null values weighted. Having weighted the risk involves in both, we decided to remove any row with null or empty value from the data, and this was done using PyCaret python library.

Data Transformation: For us to make our data to acceptable format for easy data mining and pattern recognition, it needed to undergo some data transformation. To ensure data is fully transformed to acceptable format, we used Discretization, normalization, and data aggregation technique.

### B. Implementation with Linear Support Vector Machine

Similar to SVC with parameter kernel='linear', but implemented in terms of liblinear rather than libsvm, so it has more flexibility in the choice of penalties and loss functions and scale better over a large numbers of samples. This class supports both dense and sparse input and the multiclass support is handled according to a one-vs-the-rest scheme (Supervised learning, [online] [11]. We use the inbuilt C implementation which is based on random number generator to select best features so as to ensure low bias and high variance (Fig. 5).

### C. Validation and Cross Validation

We obtain an accuracy of over 92% for both Multinomial Naïve Bayes and Linear support Vector Machine, But having such accuracy can be reason for over fitting a times. So we needed to see how the model will behave when fed or exposed to an unfamiliar data. Hence, we didn't jump to conclusion immediately; we decided to do validation and cross validation. We started our validation with confusion matrix and F1 score (Fig. 6).

Fig. 4. Python Code Snippet with Multinomial Naïve Bayes.

```
In [76]: import numpy as np
         url="C:\\Users\\data-1\\toxicity_parsed_dataset1.csv"
         twenty_test = pd.read_csv(url)
         docs_test = twenty_test.text
         predicted = text_clf.predict(docs_test)
         np.mean(predicted==twenty_test.oh_label)

Out[76]: 0.9246892011178642

In [77]: from sklearn.linear_model import SGDClassifier
         text_clf = Pipeline([('vect', CountVectorizer()), ('tfidf', TfidfTransformer()),('clf', SGDClassifier(loss='hinge', penalty='l2',

         text_clf.fit(twenty_train.text, twenty_train.oh_label)

Out[77]:
         predicted = text_clf.predict(docs_test)
         np.mean(predicted == twenty_test.oh_label)

Out[78]: 0.9239307365217711

Out[78]: Pipeline(steps=[('vect', CountVectorizer()), ('tfidf', TfidfTransformer()),
                         ('clf',
                          SGDClassifier(alpha=0.001, max_iter=5, random_state=42,
                                        tol=None))])
```

Fig. 5. Python Code Snippet with Support Vector Machine.

```
In [61]: from sklearn.metrics import classification_report, confusion_matrix, accuracy_score
         print("--------------------------------------------")
         print(confusion_matrix(twenty_test.oh_label, predicted))
         print("--------------------------------------------")
         print(classification_report(twenty_test.oh_label, predicted))
         print("--------------------------------------------")
         print(accuracy_score(twenty_test.oh_label, predicted))
         print("--------------------------------------------")

         [[28189    0]
          [ 2349  101]]
         --------------------------------------------
                       precision    recall  f1-score   support

                    0       0.92      1.00      0.96     28189
                    1       1.00      0.13      0.21      2694

             accuracy                           0.92     30883
            macro avg       0.96      0.56      0.59     30883
         weighted avg       0.93      0.92      0.90     30883

         0.9239307365217711
```

Fig. 6. Validation Report.

With an accuracy of 96% and a very low mean square error (MSE), our model is certain to have low bias and high variance, hence, we proceeded to developing a restful API using Python's flask framework and then expose our model to the API to ensure the widest coverage. We developed a chatbot automation system and exposed it to our newly developed restful API service. The chatbot automation system has a graphical user interface (GUI) for human interaction.

Since, we have exposed our restful API to our model, the API is automatically called and it immediately intercepts both outgoing and incoming messages and categorizes them based on bullying and non-bullying. If the outgoing message falls into bullying, it automatically filtered it and ensures the receiver doesn't receive it. The sender too will see from the graphical interface that the message did not deliver. In this case, the bullying message does not reach the receiver (Fig. 7).

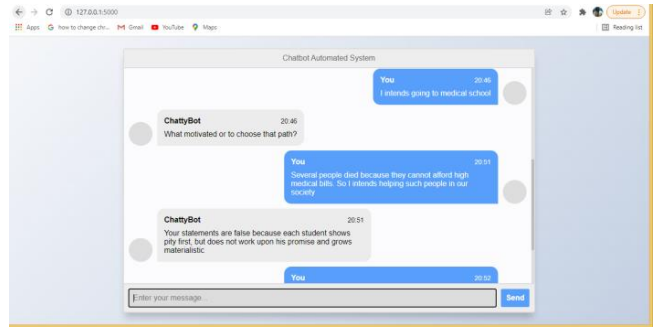

Fig. 7. Chat Messaging GUI.

### D. Additional Observation and Time Delivery Comparison

Since, there is an Artificial Intelligence Powered backend implementation that intercepts all outgoing and incoming messages, we deem it necessary to estimate the time for the message to reach the receiver and vice versa since there is an additional backend process of interception. To do this we first removed the backend interception from the newly developed chatbot automation system, and then begin to interact be sending messages from the interface. We discovered that the messages were instantly delivered. Then we decided to re-introduce our restful API which will intercept all messages and feed our trained model, we monitor if for some time during message interaction from the surface. We find out the messages was also delivered on time if it is non-bullying, but if it is bullying the message is not delivered at all. Since, there is no apparent different in time taken when we removed the backend interception process and when we added. We came to conclusion that the extra time taken by the backend message interception process which determine whether the message should be sent to the receiver or not is fraction of a seconds which is negligible.

## IV. CONCLUSION

With our research and implementation, we believed the problem of cyber bullying will be solved once and for all if our implementations can be integrated in all the online social media and messengers. Validation and cross validation of our results shows accuracy of 92%, with low bias and high variance and also a very low mean square error (MSE). Our implementation can be adopted and used on any real life social media or online messenger.

Our conclusion will be incomplete without mentioning the limitation of our developed model. The only limitation to our model is the data. The more the data used to train model, the more accuracy it will be.

The data used for this research is from different social media platforms like Kaggle, Twitter, Wikipedia Talk pages and YouTube. The data contain text and labeled as bullying or not. The data contains different types of cyber-bullying like racism, hate speech, aggression, insults and toxicity.

The greatest breakthrough in machine learning is going to be when a deployed model can automatically retrain itself by constantly getting live data from a source, train and update itself without any human effort. Since, model accuracy is somehow dependent with availability of data, and millions of new data are constantly being available in the cloud on daily bases after deploying our model. This will happen at some later time and we will be able to have a super model that is able to get live data, retrain, and redeployed itself on daily bases without human intervention.